\documentclass[letterpaper, 10 pt, conference]{ieeeconf}

\IEEEoverridecommandlockouts                              

\usepackage{cite}
\usepackage{microtype}
\usepackage{graphicx}
\usepackage{subcaption}
\usepackage{booktabs} 
\usepackage{amsmath}
\usepackage{amssymb}
\usepackage{mathtools}
\usepackage{amsfonts}
\usepackage{hhline}
\usepackage{wrapfig,lipsum,booktabs}
\graphicspath{ {./figures/} }
\usepackage{algorithm} 
\usepackage{algorithmic}
\usepackage{soul}

\usepackage{amsmath,amsfonts,bm}

\def\eqref#1{equation~\ref{#1}}

\def\1{\bm{1}}

\DeclareMathAlphabet{\mathsfit}{\encodingdefault}{\sfdefault}{m}{sl}
\SetMathAlphabet{\mathsfit}{bold}{\encodingdefault}{\sfdefault}{bx}{n}

\usepackage{cite}
\usepackage{xcolor}     
\usepackage{color, colortbl}
\definecolor{Gray}{gray}{0.9}
\definecolor{myblue}{HTML}{268BD2}
\usepackage[colorlinks,
linkcolor=myblue,
citecolor=myblue]{hyperref}
\usepackage{url}

\title{\LARGE \bf Learning Multi-Robot Coordination through\\ Locality-Based Factorized Multi-Agent Actor-Critic Algorithm\thanks{This work has been submitted to the IEEE for possible publication. 
Copyright may be transferred without notice, after which this version may no longer be accessible. 
This is the preprint version.}}

\author{
    Chak Lam Shek$^{1}$, Amrit Singh Bedi$^{2}$, Anjon Basak$^{3}$, Ellen Novoseller$^{3}$, Nick Waytowich$^{3}$, \\
    Priya Narayanan$^{3}$, Dinesh Manocha$^{4}$, Pratap Tokekar$^{4}$%
    \thanks{$^{1}$ Department of Electrical and Computer Engineering, University of Maryland, College Park, Maryland, USA}%
    \thanks{$^{2}$ Department of Computer Science, University of Central Florida, FL, USA}%
    \thanks{$^{3}$ DEVCOM US Army Research Laboratory, Adelphi, Maryland, USA}%
    \thanks{$^{4}$ Department of Computer Science, University of Maryland, College Park, Maryland, USA}%
}

\begin{document}
\maketitle

\begin{abstract}
    In this work, we present a novel cooperative multi-agent reinforcement learning method called \textbf{Loc}ality based \textbf{Fac}torized \textbf{M}ulti-Agent \textbf{A}ctor-\textbf{C}ritic (Loc-FACMAC). Existing state-of-the-art algorithms, such as FACMAC, rely on global reward information, which may not accurately reflect the quality of individual robots' actions in decentralized systems. We integrate the concept of locality into critic learning, where strongly related robots form partitions during training. Robots within the same partition have a greater impact on each other, leading to more precise policy evaluation. Additionally, we construct a dependency graph to capture the relationships between robots, facilitating the partitioning process. This approach mitigates the curse of dimensionality and prevents robots from using irrelevant information. Our method improves existing algorithms by focusing on local rewards and leveraging partition-based learning to enhance training efficiency and performance. We evaluate the performance of Loc-FACMAC in three environments: Hallway, Multi-cartpole, and Bounded-Cooperative-Navigation. We explore the impact of partition sizes on the performance and compare the result with baseline MARL algorithms such as LOMAQ, FACMAC, and QMIX. The experiments reveal that, if the locality structure is defined properly, Loc-FACMAC outperforms these baseline algorithms up to 108\%, indicating that exploiting the locality structure in the actor-critic framework improves the MARL performance.
\end{abstract}


\begin{center}
    \textbf{This work has been submitted to the IEEE for possible publication. 
    Copyright may be transferred without notice, after which this version may no longer be accessible.}
\end{center}

\section{Introduction}
\label{sec:introduction}

\begin{figure}
    \centering
    \includegraphics[width=0.8\linewidth]{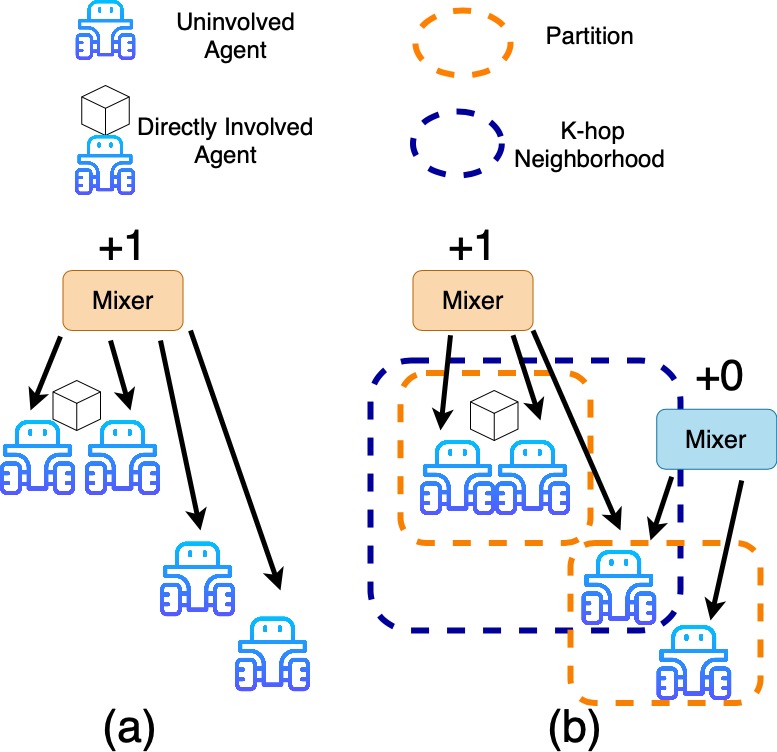}
    \caption{In a warehouse environment, the task is to deliver a package to the target location for a reward of +1 per delivery. Two robots are involved in the process while the other robots are not. (a) In the single mixer approach, all robots share the reward, even those not involved. (b) In Loc-FACMAC (ours), mixers are assigned to partitions, and only agents within the partition or k-hop neighborhood share the reward.}
    \label{fig:warehouse}
\end{figure}

Multi-Agent Reinforcement Learning (MARL) is a framework ~\cite{foerster2017counterfactual, tan1993multi} that enables a group of robots to learn team behaviors by interacting with an environment. Recently, the impact of MARL has become quite evident in a range of areas \cite{jiang2022multi,chu2019multi,zhang2022ai, he2016faster,arul2022multi}. In the field of robotics, multi-agent coordination plays a crucial role in various applications such as cooperative searching \cite{qie2019joint, vorotnikov2018multi, jimenez2018decentralized}, human-robot interaction \cite{ji2022traversing}, product delivery \cite{ota2006multi}, and soccer \cite{reis2023coordinationmachinelearningmultirobot}. In these scenarios, robots often rely on local observations to make decisions that benefit the entire team. In many MARL algorithms,  access to global rewards is assumed. However, the assumption does not hold in many scenarios as robots often need to learn cooperative behaviors based only on local observations and local or group rewards. For example, in a warehouse environment \cite{christianos2020shared, papoudakis2021benchmarking}, robots need to plan routes for product delivery in a shared space while avoiding collisions (Fig. \ref{fig:warehouse}). Existing MARL methods typically use a single global reward to train all robots. This approach fails to fully leverage the structured relationships among the robots, as a successful delivery only requires a specific subgroup of robots to cooperate in preventing congestion. In this paper, we propose a new MARL technique that learns using the locality inherent in many multi-agent coordination scenarios.

Similar to single-agent RL, most existing MARL frameworks can be classified into two categories: \textit{value-based} \cite{watkins1992q,sunehag2017value,pmlr-v80-rashid18a, rashid2020weighted,son2019qtran,kortvelesy2022qgnn,9533636} approaches and \emph{actor-critic} approaches \cite{konda1999actor,NEURIPS2021_65b9eea6, wang2020r}. In value-based approaches, robots learn to estimate an action-value function by exploring the action space and choosing the action with the maximum action value. Value-based approaches are commonly used in MARL, in part, because QMIX \cite{pmlr-v80-rashid18a} has shown the potential of solving complex coordination problems such as the Star-Craft Multi-Agent Challenge (SMAC) \cite{samvelyan2019starcraft}. The idea of QMIX has been extended in several ways, including WQMIX\cite{rashid2020weighted} and Qtran\cite{son2019qtran}.

Although value-based approaches have shown the potential for solving complicated tasks, the curse of dimensionality prevents the approach from being applied to large-scale tasks. For instance, in QMIX-type approaches, when the number of robots increases, the joint-action space exponentially increases. Meanwhile, value-based approaches must compare all action values, requiring significant search time. Therefore, value-based approaches are not an efficient method for training many robots at once.

In contrast, the actor-critic approach directly learns a policy and decides actions from the policy rather than maximizing the Q value function. For instance, MADDPG \cite{lowe2017multi} is one of the classical approaches in this category. The actor in MADDPG learns to generate the optimal action and sends the selected actions to a critic to evaluate the value of the chosen action. Then, the actor adjusts the policy according to the score given by the critic. MADDPG can reduce the training time, but agents update their policy via a separate policy gradient while assuming that the actions of all other agents are fixed. Therefore, the policy commonly falls into the sub-optimal solution. A recently proposed algorithm, FACMAC \cite{NEURIPS2021_65b9eea6}, overcomes the limitations of solely value-based and only actor-critic approaches by combining them both. In FACMAC, QMIX is used for the critic update, and a centralized policy gradient is used for the actors during training. FACMAC outperforms QMIX and MADDPG on various tasks within the SMAC environment  \cite{NEURIPS2021_65b9eea6}.

A common feature of most existing MARL approaches is that they aim to maximize a common global reward. However, in various practical applications, it is possible that one robot's actions may not have any effect on another robot. For example, in a task of surveillance by a group of robots \cite{kolling2008multi}, if two robots are quite far from each other, it makes sense to assume that their actions will not affect each other.  This idea is explored in a recent paper for value function-based MARL approaches, in which the authors proposed the LOMAQ algorithm \cite{zohar2022locality}. LOMAQ presents a multi-mixer approach to accelerate the training process by exploiting the locality of the rewards by defining a partition (a subset of agents) across the network of robots. LOMAQ provides theoretical guarantees under fully observable settings that maximizing the global joint-action value is equivalent to maximizing the action value in each partition. The partition's action value reflects the performance of the partition so each robot can learn a local policy that maximizes the local reward of that partition instead of focusing on maximizing the global reward. The feedback in each partition only updates the most correlated robots' actor-critic networks. 
\begin{table}
\caption{Comparison of different approaches. }
\centering
 \resizebox{0.45\textwidth}{!}{\begin{tabular}{cccc}
\hline
\textbf{Algorithm}        & \textbf{Num. of Mixers} & \textbf{Critic}       & \textbf{Actor}        \\ \hhline{====}
QMIX \cite{pmlr-v80-rashid18a}        & $1$        & $\checkmark$ & $\times$ \\ \hline
LOMAQ \cite{zohar2022locality}      & $K$    & $\checkmark$  & $\times$\\ \hline
FACMAC \cite{NEURIPS2021_65b9eea6}      & $1$   & $\checkmark$     & $\checkmark$ \\ \hline
\textbf{Loc-FACMAC (This work)} & \bm{$K$}   & \bm{$\checkmark$} & \bm{$\checkmark$} \\ \hline
\end{tabular}}
\label{table_one}
\end{table}

Several works have examined the relationship between robots \cite{hao2023boosting}. Recent approaches utilize attention mechanisms to determine the weights of graph neural networks (GNNs) \cite{liu2019multiagent, li2021deep}, which connect the robots' actions to cooperative behavior. Alternatively,  deep coordination graphs (DCG) \cite{wang2022contextaware, böhmer2020deep} can be used to model the payoffs between pairs of robots. However, existing works dynamically learn the graph during end-to-end learning, resulting in continuous changes to the graph structure.

In this work, we extend the concept of \emph{locality} to actor-critic methods and introduce a novel locality-based actor-critic approach named \textbf{Loc}ality-based \textbf{Fac}torized \textbf{M}ulti-Agent \textbf{A}ctor-\textbf{C}ritic Algorithm (Loc-FACMAC) for cooperative MARL. The key characteristics of Loc-FACMAC compared to existing methods are summarized in Table \ref{table_one}. Loc-FACMAC separates the process of constructing the dependency graph from policy learning. Once the dependency graph is established, Loc-FACMAC utilizes it across multiple mixers to compute the local joint-action value in each partition. Both critic and actor can leverage locality information to update the network with accurate policy gradient values. Similar to FACMAC, Loc-FACMAC divides the training process of critics and actors, enabling the critic to precisely evaluate action value quality without being influenced by action choices. The actor learns from the high-quality local action-value function and rapidly converges to an optimal policy.

\textbf{Contributions.} We summarize our {main contributions} as follows. 
\begin{itemize}
  \item We introduce a novel two-stage MARL approach named Loc-FACMAC. In the first stage, we construct a dependency graph based on the variance of agents' performance when they utilize state and action information from other agents. In the second stage, agents learn the local policy within the fixed structure of the dependency graph using an actor-critic approach. 
    
  \item We evaluate the proposed Loc-FACMAC algorithm in three MARL environments: Multi-cartpole, Bounded-Cooperative-Navigation, and Hallway. Our framework achieves strong performance, showing a 108\% improvement over baseline methods in solving these tasks. If a proper dependency graph is defined, our framework can achieve the maximum reward with a competitively short amount of training time.

  \item We also examine the Loc-FACMAC in different partitions and analyze the agent relationships using dependency graphes. The analysis shows that there is an improvement when partitions are optimized.

\end {itemize}

\begin{figure}
    \centering
    \includegraphics[scale=0.24]{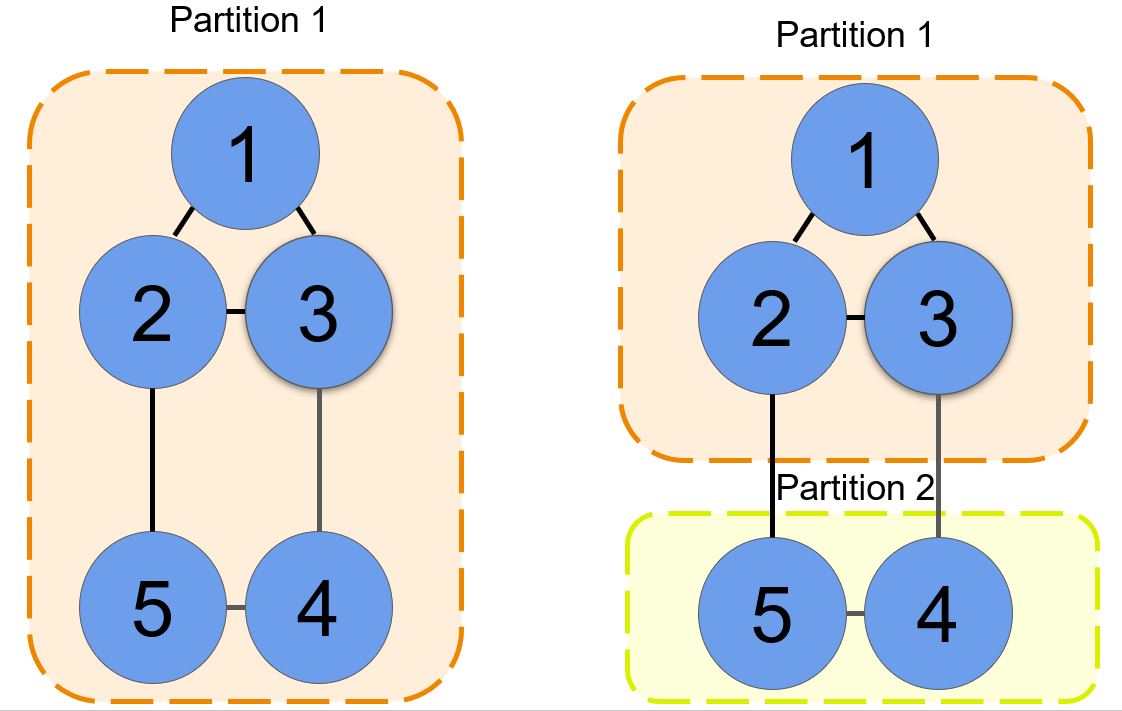}
    \caption{This figure considers a five-robot network and shows how different robots are connected. It also shows two ways of partitioning the network to leverage locality in the learning process. Robot 4 and robot 5 are away from robot 1, robot 2, and robot 3. Robot 4 and robot 5 can be grouped separately.}
    \label{fig:my_label0}
\end{figure}
%


\begin{figure}[ht]
    \centering
    \includegraphics[width=0.5 \textwidth]{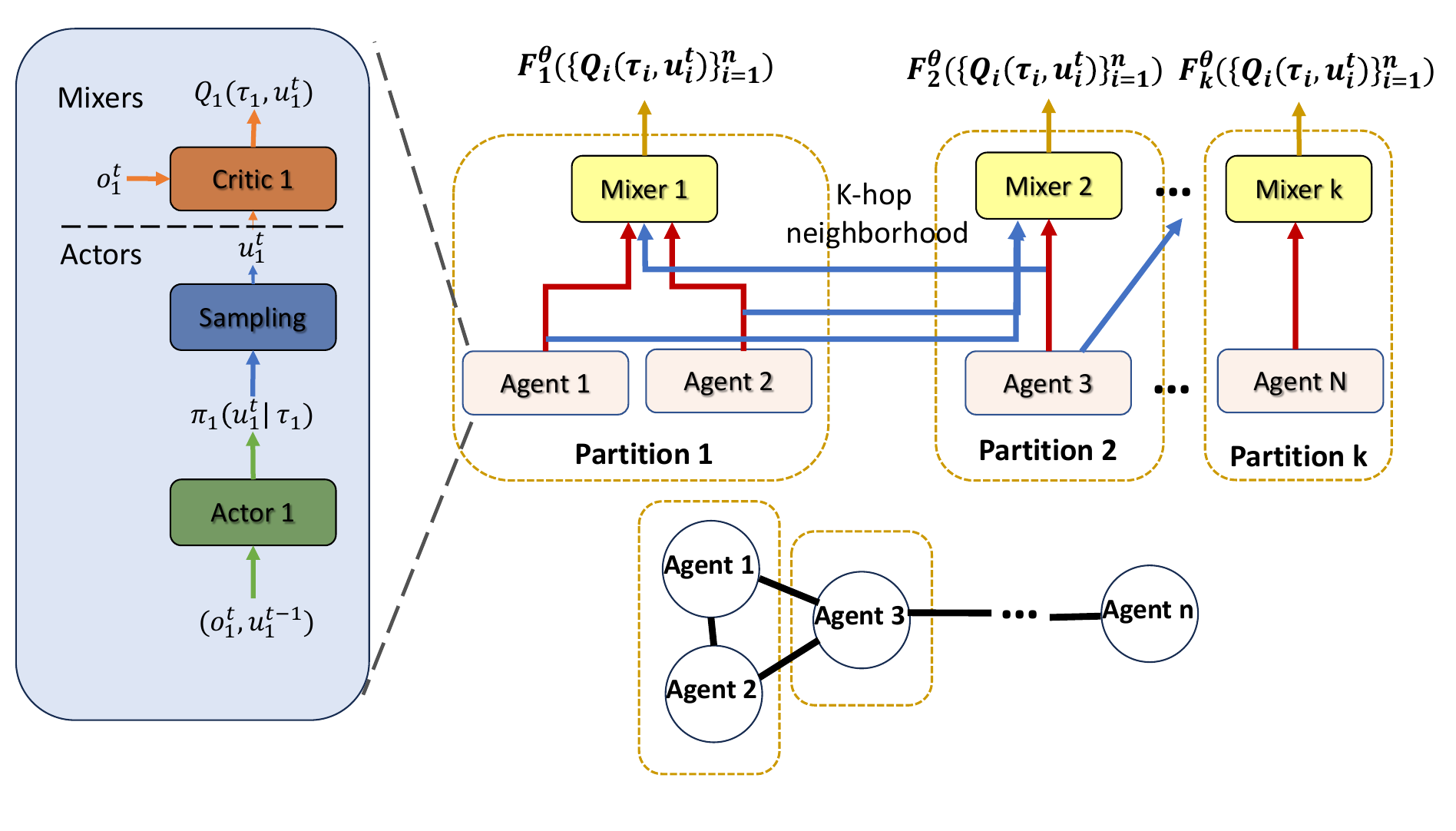}
    
    \caption{This figure presents the architecture of the proposed Loc-FACMAC. Our proposed framework, Loc-FACMAC, consists of Actors, Critics, and Mixers.} 
    \label{fig:algorithm}
\end{figure}

\section{Problem Formulation}
\label{sec:problem}
In this paper, we focus on coordinating robots to complete cooperative tasks. The robots are grouped into partitions and can observe the rewards specific to their respective partitions. We model the problem as a decentralized, partially observable Markov decision process (Dec-POMDP).  We define the process as a tuple $\mathcal{M} = (N , \mathcal{S}, \mathcal{A}, P, r, \Omega, O, \gamma, \mathcal{G})$. Here, $N$ denotes a finite set of robots,  $s \in \mathcal{S}$ is the true joint state, and $\mathcal{A}$ is the joint-action space of all robots. At each time step, if the state is $s \in \mathcal{S}$, and each robot $i$ selects an action from a continuous or discrete action space $\mathcal{A}_i$, then we transition to state $s'\sim P(\cdot|s, a) $, where  $P(\cdot|s, a)$ represents the transition kernel from state $s$ to $s'$. We define $r$ as the global reward which depends on the global state and the joint action.   The discount factor is denoted by $\gamma$. We note that $\Omega$ is the observation space, which implies that at each time step, robot $i$ can observe partial information $o_i$ sampled from $O_i(s, a) \in \Omega$.

$\mathcal{G} = (\mathcal{V}, \mathcal{E})$ is an undirected graph of robots, namely a dependency graph, where $\mathcal{V} =\{1,2,...,N\}$ denotes the node and $\mathcal{E} \subseteq \mathcal{V} \times \mathcal{V}$ is the edge between nodes. The dependency graph is an additional feature added to the original Dec-POMDP in this work. The two robots are correlated if a robot is connected with another robot in the dependency graph, as shown in Fig. \ref{fig:my_label0}. Hence, the action of a robot affects its neighbor robot's reward. The goal here is to learn a stochastic policy $\pi_i(a_i|\tau_i)$ or a deterministic policy $\mu_i(\tau_i)$ for each robot $i$ where $\tau_i$ is the  local action-observation history $\tau_i \in \mathcal{T} = \Omega \times \mathcal{A}$.

{We assume that the dependency graph can be decomposed into a collection of partition $\mathcal{P} =\{J_k\}_{k=1}^K$, such that $J_k \cap J_l = \emptyset, \forall k \neq l$ and $\bigcup_{k} J_k=\mathcal{V}$. The global reward $r$ is expressed by $\{r_1, r_2,...,r_K\}$, such that $r= \sum_{k=1}^K r_k$.}

\begin{algorithm}[ht]
\caption{{Proposed Loc-FACMAC Algorithm}}
\label{alg:cap}
\begin{algorithmic}[1]
\ENSURE $i \in {N} = 1,2,..., n$ are the robots
\STATE Partition $\mathcal{P}$ of $V$
\STATE Policy $\pi^{i}_{\theta_i}(\tau_i) $ with parameter $\theta_i$, where $\tau_i \in \Gamma \equiv (\Omega \times A)$ is local action-observation history

\STATE Local state $s^i_0 \in \Omega$ drawn from $O(s,i)$, where s is true state of environment

\STATE Critic Networks $\phi_i \in \mathbb{R}^d$

\STATE Mixing networks $\psi_k \in \mathbb{R}^{e}$, where $k = 1,2,.., K$

\FOR{iter $t = 1,2, ...,m$}
    \FOR{robot $ i = 1,2,..., n$}
        \STATE Sample action $a_i(t)$ from $\pi^i_{\theta_i}(\tau_i)$ and retrieve next observation $s(t + 1)$ and reward $r(t)$
    \ENDFOR
    \STATE We have $Q_J^\pi(\boldsymbol{\tau}^t, \boldsymbol{a}^t,\boldsymbol{s}^t;\boldsymbol{\phi}^t_N, \boldsymbol{\psi}^t_J)= F^t_{\psi_{J}}(s_t,\{Q^{\pi_i}_i(\tau^t_i,a^t_i;{\phi}^t_i)\}_{i \in \mathcal{N(J)}})$, $\mathcal{N(J)}$ is the $\kappa$-hop neighbourhood of $J$ in the dependency graph

    \STATE
    \STATE \textbf{Train mixing and Critic network:} {update mixing network by minimizing the loss:}
    
    \STATE
    $ L(\boldsymbol{\phi},\boldsymbol{\psi}) = E_D[\sum_{J \in \mathcal{P}} (y_J^{tot} - Q_J^\pi(\boldsymbol{\tau}^t, \boldsymbol{a}^t,\boldsymbol{s}^t;\boldsymbol{\phi}^t_N, \boldsymbol{\psi}^t_J))^2] $
    
    \STATE ${\phi}_i \leftarrow {\phi}_i -	\alpha \nabla_{\phi_i} L(\boldsymbol{\phi},\boldsymbol{\psi})$
    \STATE $\psi_k \leftarrow \psi_k -	\beta \nabla_{\psi_k}   L(\boldsymbol{\phi},\boldsymbol{\psi}) $

    \STATE
    \STATE \textbf{Train Actor network:} by update
    \STATE $\nabla_\theta J(\theta) = E_D[\nabla_\theta \pi \nabla_\pi Q^\pi_{tot}(\boldsymbol{\tau}^t, \pi_1(\tau_1^t), ... \pi_n(\tau_n^t))]$
    \STATE Update policy parameter $\theta_k \leftarrow \theta_k -	\gamma \nabla_\theta   J(\theta) $
    
\ENDFOR
\end{algorithmic}
\end{algorithm}

We also define the global action-value function as 
\begin{align}
Q_{tot}(\boldsymbol{\tau}(t), \boldsymbol{a}(t))= \mathbb{E}\bigg[\sum_{t=0}^\infty \gamma^t r(\boldsymbol{\tau}(t),\boldsymbol{a}(t))\bigg],
\end{align}
where we have $\boldsymbol{\tau}(0)=\boldsymbol{\tau}(t),\boldsymbol{a}(0)=\boldsymbol{a}(t)$. 
Similarly, we define the local Q function $Q_i (\tau_i(t), a_i(t))= \mathbb{E}[\sum_{l=0}^\infty \gamma^t r_i(\tau_i(t),a_i(t))],$ with $\tau_i(0)=\tau_i(t),a_i(0)=a_i(t)$. We assume the condition $Q_{tot}(s(t),a(t))= \sum_{i=1}^N Q_i(s_i(t),a_i(t))$ holds at any time step which is {well-defined} in value function based approaches \cite{rashid2020weighted,zohar2022locality}.

\section{Proposed Algorithm: Loc-FACMAC}

\subsection{Algorithm Design}

In this section, we present the framework of our new algorithm \ref{fig:algorithm}, Loc-FACMAC, which is a multi-mixer actor-critic method that allows the robots to synchronize the policy update while the locality information of the rewards is retained. 

Loc-FACMAC is built upon FACMAC and overcomes FACMAC's limitation of overgeneralized policy gradients. In FACMAC, the robots' network is updated using the policy gradient computed from the global reward \cite{NEURIPS2021_65b9eea6}.  Even though the robot does not contribute to the global reward, it still updates its network using the same policy gradient. Using the distorted policy gradient, robots fail to estimate the action value and end up with a sub-optimal policy. Therefore, we adopt the idea of the locality of rewards. The policy gradient is computed from the partitions' reward and used to update the strongly related robots' network. Compared to the policy gradient in FACMAC, the policy gradient computed from partitions' reward is more accurate, so it can give a better update direction to the policy network. 

In the Loc-FACMAC, multiple mixers are used to estimate the locality of rewards. Each mixer corresponds to one partition. It takes the local action values and maps them to the partition's joint-action value. By increasing the number of mixers, detailed information on the locality of rewards can be retained. A simple example is the estimation four natural numbers $(n1,n2,n3,n4)$ that sum up to a given number. If only the total sum is given, the number of possible combinations could be very large. However, if the subset sums are given, e.g., $n1+n2$ and $n3+n4$, the number of possible combinations is reduced. Therefore, the extra information on the locality of rewards can aid robots in finding the globally optimal policy. Then, the loss can be calculated by summing the error in each partition using,
\begin{equation}
	  L(\boldsymbol{\phi},\boldsymbol{\psi}) = E_D[\sum_{J \in \mathcal{P}} (y_J^{tot} - Q_J^\pi(\boldsymbol{\tau}^t, \boldsymbol{a}^t,\boldsymbol{s}^t;\boldsymbol{\phi}^t_N, \boldsymbol{\psi}^t_J))^2]
\end{equation}

where $y_{J}^{tot}=\sum_{j \in J} r_j  + \gamma  \max_{a^{t+1}}  F_{\psi_J}^{t+1} (s_{t+1} , \{Q_{i}( \tau^{t+1}_i, a^{t+1}_i)\}_{i \in \mathcal{N(J)}})$ is the target reward of partition $J$. $\mathcal{N(J)}$ is the $\kappa$-hop neighbourhood of $J$ in the dependency graph.

Loc-FACMAC also inherits the property of FACMAC that it separates the learning process of actor and critic. The framework of Loc-FACMAC consists of two parts: Actor and Mixer. The actor generates an optimal action based on the local observation and historical actions and the mixer criticizes the performance of the robots' action and computes the loss of the network. The advantage of separating the learning process is to prevent the actor from learning from the overestimated/underestimated temporal difference 
(TD) error. This is critical in MARL because a robot should learn optimal policy by considering the other robots' chosen actions during updating the policy network. However, policy gradient updates the robot's policy network by assuming the other robots' policy is fixed. The assumption fails to be maintained. Especially, when the number of robots increases, the dynamic of robots' behavior becomes hard to predict. Hence, robots frequently get trapped  in a sub-optimal policy using MADDPG or QMIX, since no individual robot can modify its optimal action conditioning by fixing all other robots' actions under the sub-optimal policy. This problem is resolved by actor-critic methods in that the mixer first measures the true error of the estimated reward. Then, the actor can be updated by computing the policy gradient from the global joint-action. Since the mixer and actors are updated in two independent steps, robots can synchronize the actor update by considering other robots' actions. The policy gradient is 
\begin{align}
	\nabla _{\theta} J(\theta) = E_D [\nabla _{\theta} \mathbf{\pi} Q_{tot}^\pi (\boldsymbol{\tau}^t, \pi_1^t(\tau_1, \theta_1), ...,\pi_n^t(\tau_n, \theta_n)))] 
\end{align}
where $\pi={\pi_1^t(\tau_1, \theta_1), \pi_2^t(\tau_2, \theta_2),...,\pi_n^t(\tau_n, \theta_n))}$ is the collection of all robots' current policy. Using (3), the policy update of a robot does not only rely on its local observation and local action. Instead, the policy gradient is computed using the lasted estimated error from the mixer and the global sampled action $\mu$. 

The details of the Loc-FACMAC framework are shown in Algorithm \ref{alg:cap}. Each robot has its individual actor and critic. The actor takes the inputs of current observation $o_i^t$ and previous action $\mu_i^{t-1}$ to compute conditional policy $\pi_i$. Then, the current action $\mu_i^t$ can be sampled from the policy $\pi_i$ and fed into the critic. The critic produces the utility function $U_i$ which evaluates actions taken by the actor. In the last stage, there are $K$ mixers. Each mixer takes the outputs of the critic and the global state to find a monotonic function $F$ estimating the total rewards of a partition $i$. The number of mixers is equal to the number of partitions $\mathcal{P}$. Each partition contains a subgroup of robots and each robot has to belong to one partition, such that $J_i \cap J_j = \emptyset, \forall i,j \in 1,...,k$. For example, the set of Partition 2 in Figure 1  is denoted as $J_2 = \{4,5\}$. If $\kappa = 1$ is selected, the mixer takes the output of \{2, 3, 4, 5\} robots' critic as input to estimate $Q_{J_2}$.

\subsection{Construct Dependency Graph} \label{sec:CDG}
The Loc-FACMAC algorithm presented in the previous section assumes that the partitions are already given. In this section, we show how to construct a dependency graph and the partitions.

Before training for the main policy, we first conduct a pretraining process to construct the dependency graph. The dependency graph, which delineates the relationship between robots' actions, is an important component of Loc-FACMAC. The topology of the dependency graph influences both the training efficiency and the performance of the models. An optimal dependency graph manifests as a sparse structure,
connecting only a select few robots of the most significance. Hence, building on the insights of \cite{wang2022contextaware}, we quantify the influence between robots, defined as the discrepancy in robot \(i\) with the inclusion of information from robot \(j\). This is formalized as:
\[
\xi _{ij} = \max_{a_i} \text{Var}_{a_j} \left[ q_i^j(\tau_i,\tau_j, a_i,a_j)) \right]
\]
where \(\text{Var}\) represents the variance function, capturing the uncertainty or variability in robot \(i\)'s value function due to the actions of robot \(j\). The term \(q_i^j\) refers to the Q function of robot \(i\) considering the joint state and action information from both \(i\) and \(j\).

To estimate \(q_i^j\), we utilize Independent Q-learning to learn individual value functions for each robot in algorithm \ref{alg:new_algo2}. However, the methodology for constructing the dependency graph is not limited to Independent Q-learning. Other learning frameworks, such as actor-critic approach, can also be applied by adjusting the estimation of \(q_i^j\). This flexibility allows the algorithm to generalize across different MARL architectures while maintaining its core objective of identifying critical inter-robot dependencies for efficient communication and coordination.

\begin{algorithm}
\caption{{Algorithm for Constructing the Dependency Graph}}
\begin{algorithmic}[1]
\label{alg:new_algo2}
\ENSURE $i \in N  = 1,2,..., n$ are the robots
\STATE Initialize replay buffer $D$
\STATE Initialize empty graph $G$
\STATE Set threshold $\xi_{th}$

\FOR{robot $i = 1,2,...,n$}
    \FOR{each robot $j = 1,2,...,n, j \neq i$}
        \FOR{iter $t = 1, ..., m$}
            \STATE Generate samples from environment and store them in buffer $D$
            \STATE Update $Q_k(\tau_i, a_i)$ for all robots $k \neq i$
            \STATE Update $Q_k(\tau_i, \tau_j, a_i, a_j)$ for $k = i$ 
        \ENDFOR
        \STATE Compute $\xi_{ij} = \max_{a_i} \text{Var}_{a_j} \left[ q_i^j(\tau_i, \tau_j, a_i, a_j) \right]$
        \IF{$\xi_{ij} > \xi_{th}$}
            \STATE Connect edge $(i, j)$ in graph $G$
        \ENDIF
    \ENDFOR
\ENDFOR

\RETURN Graph $G$
\end{algorithmic}
\end{algorithm}

\begin{table}[!ht]
\caption{Experiment Hyperparameters}
\begin{tabular}{p{0.2\textwidth}p{0.24\textwidth}} 
\hline
\textbf{Hyperparameter} & \textbf{Hallway/Multi-Cartpole/BCN}                                                    \\ \hhline{==}
$\epsilon_{start}$                & 1.0 / 1.0 / 1.0                                        \\ \hline
$\epsilon_{end}$                  & 0.05 / 0.05 / 0.05                                                   \\ \hline
$\epsilon$ anneal time            & 100000 / 20000 / 100000                                           \\ \hline
batch size                        & 64 / 128 / 128                                               \\ \hline
$\gamma$                          & 0.99 / 0.99 / 0.99                                \\ \hline
$\kappa$                            & 1 / 1 / 1                                                      \\ \hline  
actor learning rate               & 0.0025 / 0.0025 / 0.0025                                              \\ \hline
mixer and critic learning rate    & 0.0005 / 0.008 / 0.0005                                                   \\ \hline
td lambda                         & 0.2 / 0.2 / 0.2                                                      \\ \hline
target update mode                & soft / soft / soft                                                    \\ \hline
target update rate                & 0.5 / 0.5 / 0.5                                                   \\ \hline
\label{table 1}

\end{tabular}
\end{table}

\section{Experiments}
\begin{figure}
    \centering
    \includegraphics[scale=0.25]{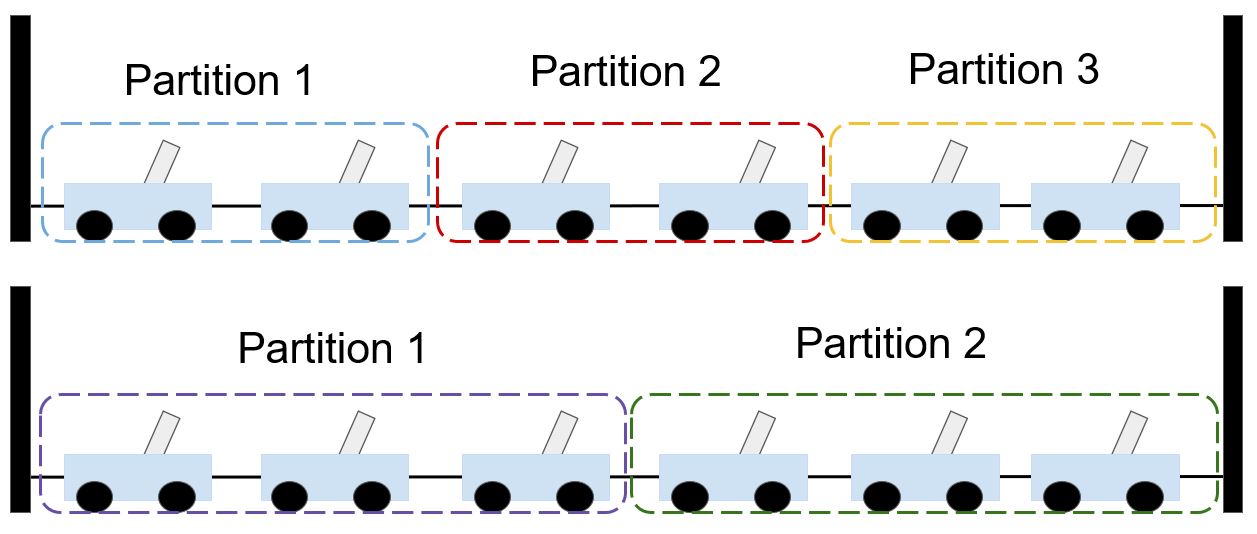}
    \caption{This figure describes the partition for one specific environment of multi-cartpole. We can divide six carts into two possible partitions (2-2-2 and 3-3). There exist other possible partitions, such as 1-2-3. }
    \label{fig:my_label_cart}
\end{figure}

\begin{figure}
    \centering
    \includegraphics[width=0.8\linewidth]{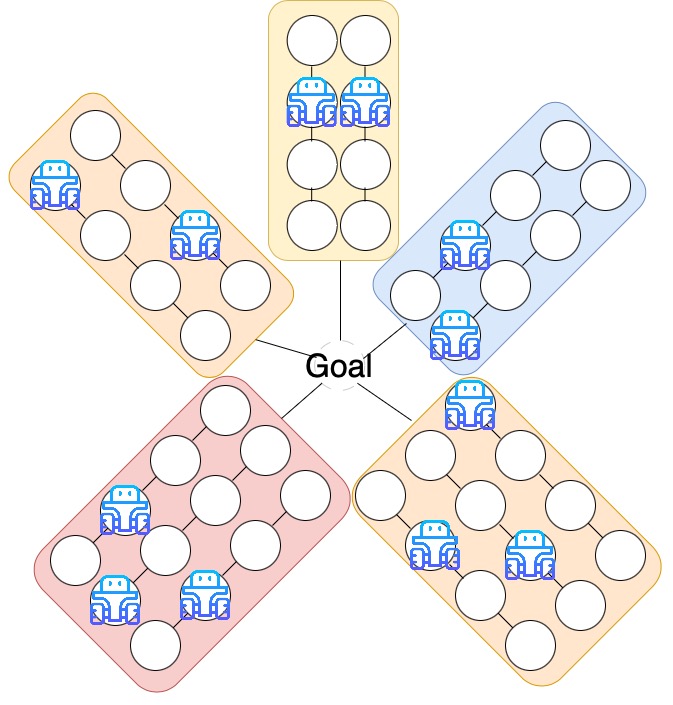}
    \caption{This figure describes the partition for one specific
environment of Hallway. We can divide twelve robots into
five partitions (2-2-2-3-3).}
    \label{fig:my_label_Hallway}
\end{figure}

\begin{figure}
    \centering
    \includegraphics[width=0.5\linewidth]{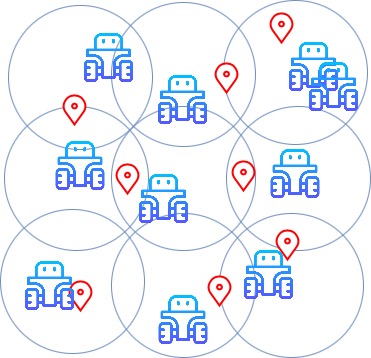}
    \caption{This figure describes a 9 robots Bounded-Cooperative-Navigation Environment.}
    \label{fig:my_label_BNC}
\end{figure}

In the experiment section, we specifically selected three distinct environments—Coupled Multi-Cart-Pole, Bounded Cooperative Navigation, and Hallway—where teams of robots need to cooperatively carry out a task to evaluate the performance of Loc-FACMAC. The Coupled Multi-Cart-Pole environment, a multi-robot version of the standard cartpole robot (Figure \ref{fig:my_label_cart}), simulates multiple robots working together to uphold objects for delivery tasks. Both the Bounded Cooperative Navigation (Figure \ref{fig:my_label_Hallway}) and Hallway environments (Figure \ref{fig:my_label_BNC}) represent warehouse settings where robots must deliver objects to specific goals while simultaneously avoiding collisions and conflicts of interest.

The selected environments were carefully chosen to cover a broad range of scenarios, emphasizing different types of state spaces, reward structures, and dependency graph structures. This selection highlights Loc-FACMAC's flexibility and adaptability in addressing diverse challenges.

In terms of state space, both Bounded Cooperative Navigation and the Multi-Cart-Pole environments feature continuous state spaces, while the Hallway environment is based on a structured, graph-like state space. This distinction demonstrates Loc-FACMAC's capability to function effectively in both continuous and discrete settings.\\

The reward structures also vary across these environments. In the Hallway environment, rewards are shared among smaller groups of robots, reflecting tasks that require localized cooperation for success. In contrast, Bounded Cooperative Navigation and Multi-Cart-Pole provide individual rewards to each robot based on their own actions, better-representing scenarios where independent contributions drive task completion.

Additionally, the dependency graph in Multi-Cart-Pole is linear. This contrasts with the more complex dependency structures found in Bounded Cooperative Navigation and Hallway, where interactions between robots are less linear and require dynamic coordination among multiple agents.


\subsection{Evaluation Environments}

\subsubsection{Hallway}
In Hallway, shown in Figure \ref{fig:my_label_Hallway}, each robot starts at a random position within a predefined chain, and the objective is for all robots within each partition to reach a common goal state simultaneously. Each partition forms a distinct group of robots. The robots have three possible actions: moving left, moving right, or staying still. The reward function awards a value of +1 if all robots in a partition reach the goal at the same time. If any robot arrives early, the entire group is disqualified from receiving a reward, and if multiple groups attempt to reach the goal simultaneously, they are prevented from moving forward. The dependency graph of the environment can be represented as a fully connected graph within each partition. In addition, each agent in one partition is connected to at least one agent in another partition, allowing for influence between the groups. To prevent stalling, the environment imposes a time limit of 300 timesteps for completing the task. The robots are divided into five partitions: $ P_i = \{\{1,2\}, \{3,4\}, \{5,6\}, \{7,8,9\}, \{10,11,12\}\} $, where each partition operates as a separate team.

\subsubsection{Coupled-Multi-Cart-Pole}
The Coupled-Multi-Cart-Pole problem, depicted in Figure \ref{fig:my_label_cart}, is based on the standard Cart-Pole problem and was developed by Zohar et al. \cite{zohar2022locality}. The objective is for each cart to maintain its pole in an upright position, earning a reward of +1 for every timestep the pole remains balanced. The carts are linked to their neighboring carts by springs, such that the movement of one cart affects its neighboring carts, creating an interdependent system. This coupling can be represented as a linear graph, where each cart's motion only impacts its adjacent neighbors. In the simulation, each robot tries to keep its respective poles upright over 300 timesteps. Since the rewards are individual and each cart earns rewards independently, a natural partition is to assign each cart to its own partition, $P_i = \{i\}, i = 1,2,...,6$, representing the maximum partition configuration.

\subsubsection{Bounded-Cooperative-Navigation Environment}\label{sec:BCN}
In the Bounded-Cooperative Navigation task (Figure \ref{fig:my_label_BNC}), the locations of nine robots are restricted within nine bounded regions, allowing them to move freely within their designated areas. These bounded regions may overlap, enabling multiple robots to access the same space. The objective for the robots is to cooperatively cover all landmarks, earning a reward of +1 for each landmark successfully covered. The behavior of each robot influences others that share a common bounded area; therefore, if any two robots are in an overlapping region, they are connected in the dependency graph. The time limit for the task is set to 50 timesteps. We applied Loc-FACMAC to the Bounded-Cooperative Navigation task with nine robots, with each robot assigned to its own partition, represented as $ P_i = \{i\}, i = 1,2,\ldots,9 $.

\begin{figure}
    \centering
    \includegraphics[scale=0.5]{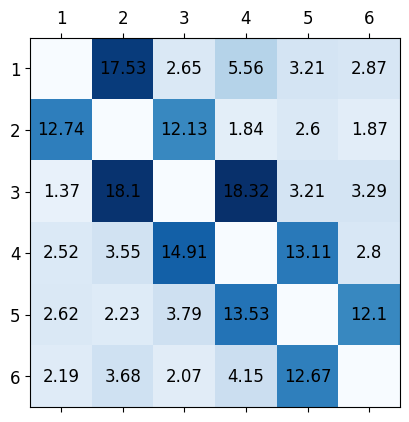}
    \caption{The measurement $\xi_{ij}$ for all pairs of robot in Coupled-Multi-Cart-Pole environment. The higher value indicates a strong relationship between the robots.}
    \label{ReL}
\end{figure}
\begin{figure}
    \centering
    \includegraphics[scale=0.4]{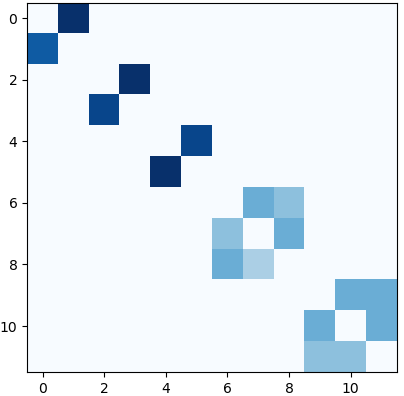}
    \caption{The measurement $\xi_{ij}$ for all pairs of robot in Hallway environment.}
    \label{ReL2}
\end{figure}
\begin{figure*}[ht]
    \centering
    \subfloat[]{
    \includegraphics[width=0.32 \textwidth]{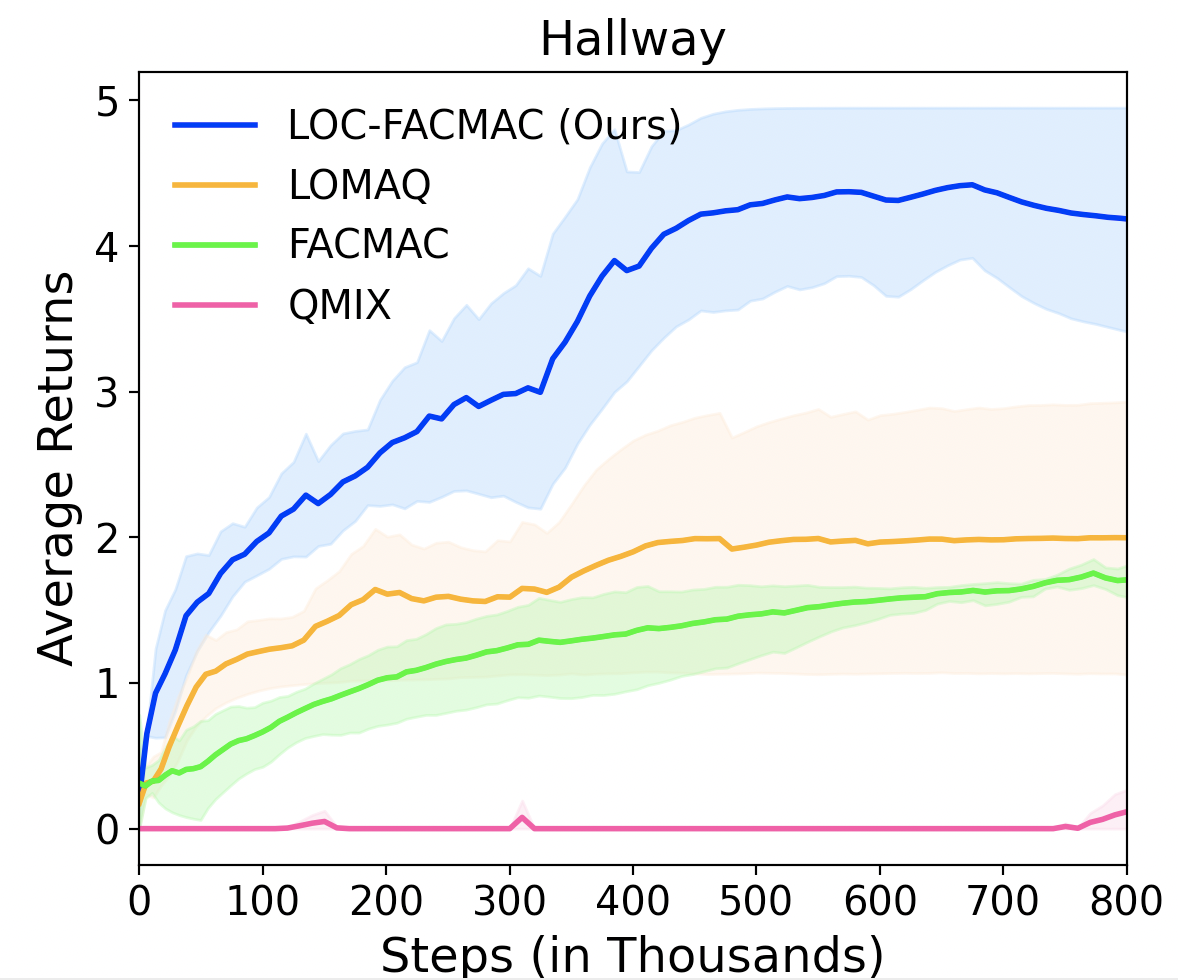}
    \label{fig:Hallway}
    }
     \subfloat[]{
    \includegraphics[width=0.34 \textwidth]{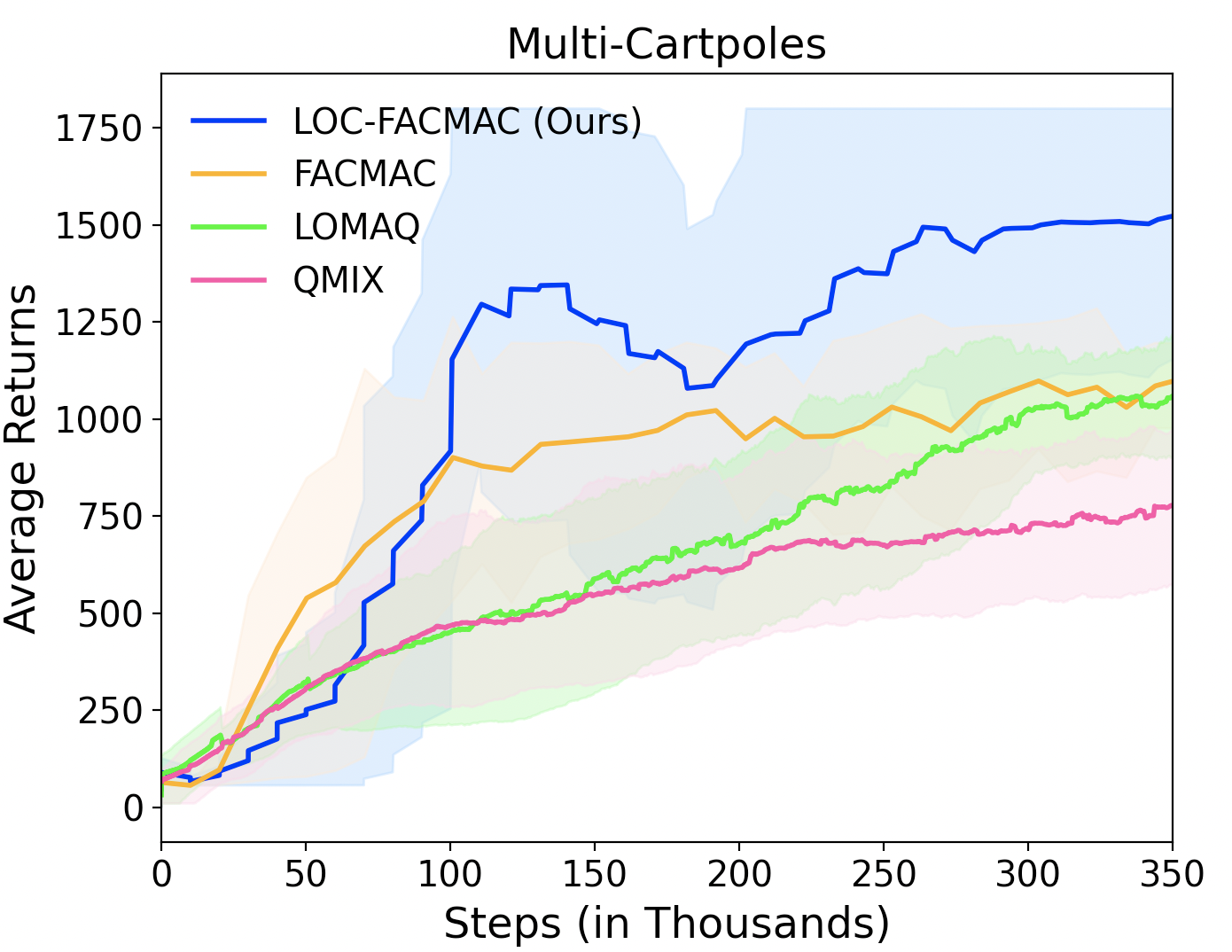}
    \label{fig:my_label1}
    }
    \subfloat[]{
    \includegraphics[width=0.32 \textwidth]{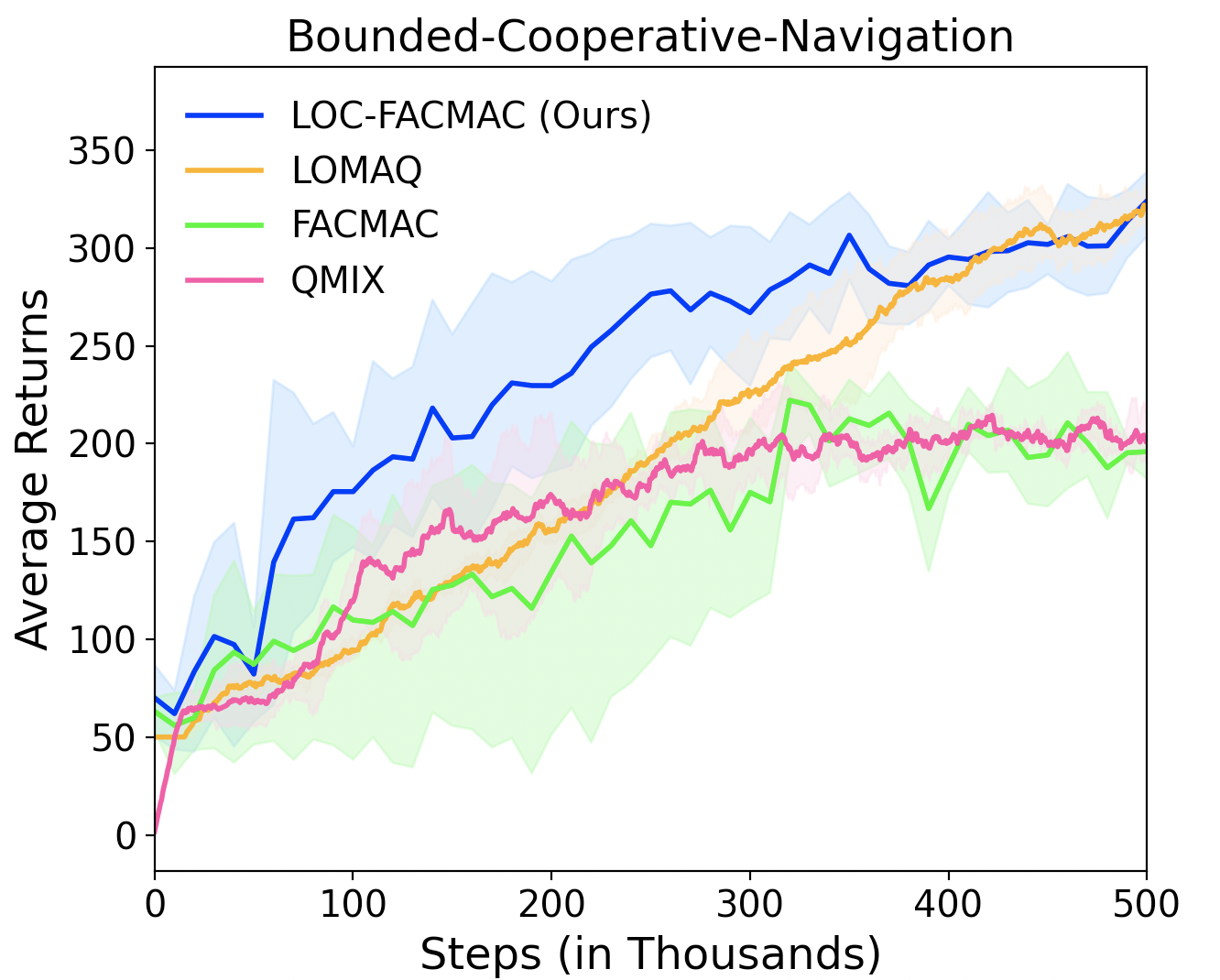}
    \label{fig:my_label3}
    }
    \caption{This figure compares the performance of the proposed loc-FACMAC algorithm against the baseline algorithms QMIX, LOMAQ, and FACMAC. }
    \label{fig:three graphs}    
\end{figure*}

\subsection{Loc-FACMAC Implementation}\label{appendix_hyper}

Most of our training hyperparameters for Loc-FACMAC are provided in Table \ref{table 1}. The architecture of the actor network consists of a linear layer, followed by a ReLU activation, a GRU with a hidden state dimension of 64, and another linear layer. The critic network is structured with three linear layers, each with an embedding dimension of 64. The mixer network is composed of alternating linear and ReLU layers, specifically: Linear, ReLU, Linear, ReLU, and Linear, with a mixing embedding dimension of 32. Results are averaged over 5 seeds. Parameter sharing is not applied in the mixer.

\subsection{Dependency Graph Qualitative Example}

We first present results on constructing a dependency graph (Section \ref{sec:CDG}). Recall that we use $\xi_{ij}$ to encode the discrepancy in robot $i$'s learning process using information from robot $j$. Figure \ref{ReL} illustrates the $\xi_{ij}$ values for all pairs of robots in the Multi-Cart-Pole environment. To derive these $\xi_{ij}$ values, we employ independent Q-learning.

Our findings align with our intuition for the Multi-Cart-Pole, suggesting that the dependency graph should be linear. We observe that the $\xi_{ij}$ values are high for immediate neighbors and low for others. Therefore, for an appropriate threshold, we obtain a linear dependency graph that can confirm our initial hypothesis.

In addition, we analyze the Hallway environment, as depicted in Figure \ref{ReL2}. The $\xi_{ij}$ values for the Hallway environment similarly reflect our expectations of the dependency graph structure. We anticipate clear connections in the Hallway, indicating that agents within the same partition group have a stronger influence on each other compared to agents in other groups. This consistency across both environments strengthens the reliability of our dependency graph construction.

\begin{figure}
    \centering
    \includegraphics[width= 0.33\textwidth]{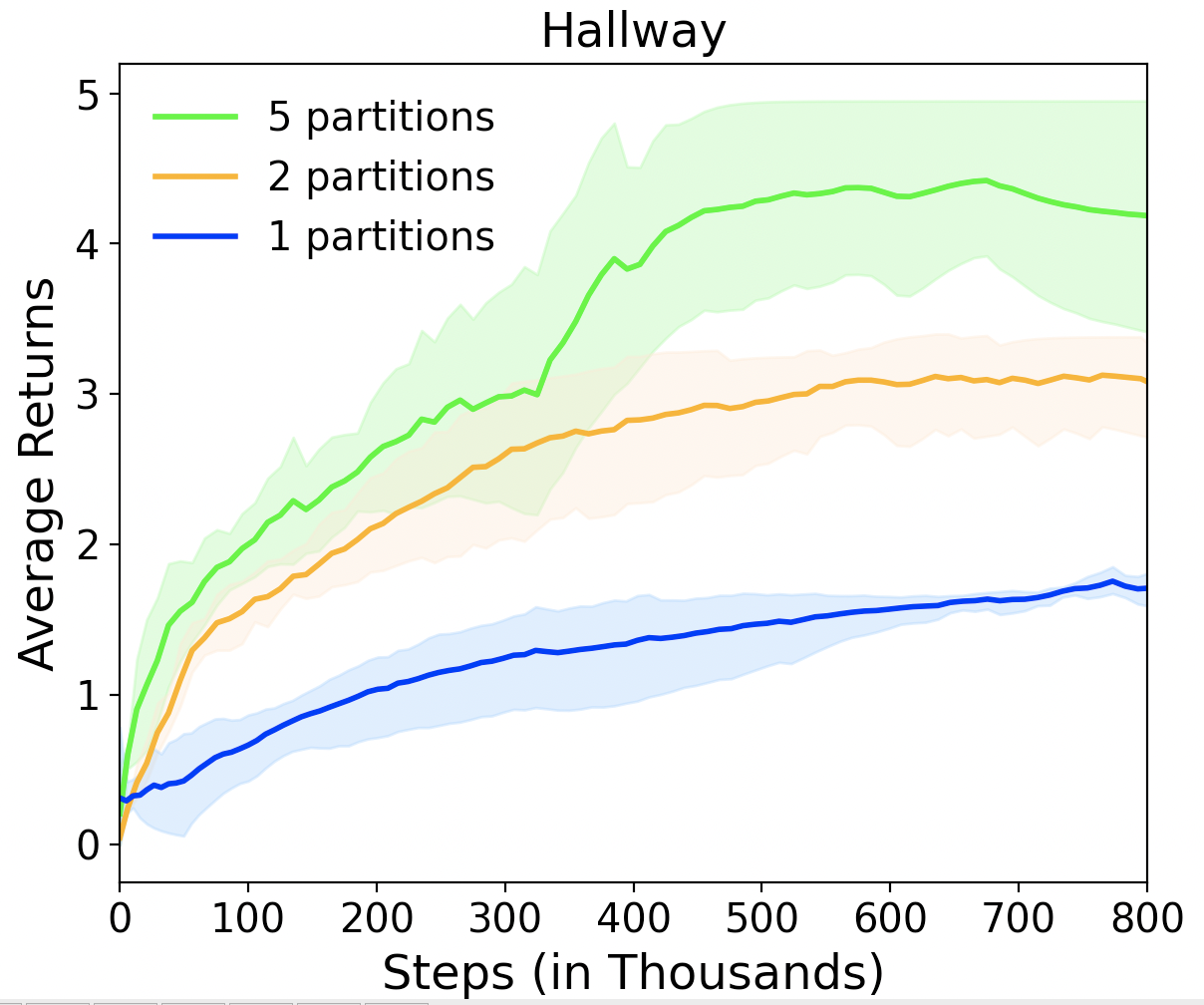}

    \caption{The simulation compares different partitions for one specific environment of the Hallway using Loc-FACMAC.}
    \label{fig:PP}
\end{figure}

\subsection{Quantitative comparisons}
In this section, we demonstrate the advantages of our proposed algorithm by comparing it with the other state-of-art MARL algorithms, QMIX \cite{pmlr-v80-rashid18a}, FACMAC \cite{NEURIPS2021_65b9eea6}, and LOMAQ \cite{zohar2022locality} across three discrete cooperative multi-robot tasks: Hallway, Coupled-Multi-Cart-Pole, and Bounded-Cooperative Navigation. As shown in Figures \ref{fig:Hallway}, \ref{fig:my_label1}, and \ref{fig:my_label3}, our algorithm consistently outperforms the others in all environments, with FACMAC and LOMAQ exhibiting moderate performance while QMIX performs the worst.

The superior performance of our Loc-FACMAC algorithm can be attributed to its unique integration of the strengths of both LOMAQ and FACMAC. Specifically, Loc-FACMAC leverages multiple mixers to effectively exploit locality, similar to LOMAQ, while incorporating the actor-critic method from FACMAC to ensure faster convergence. This combination allows Loc-FACMAC to overcome the limitations of LOMAQ, which lacks an actor, and FACMAC, which relies on a single mixer.

In the Bounded-Cooperative Navigation task, we observed that QMIX and FACMAC perform comparably, despite their lower overall rewards. This suggests that while these algorithms capture some locality, they struggle to accurately estimate local rewards from global computations. In contrast, Loc-FACMAC effectively synchronizes policy updates and explores the optimal policy more quickly. It leads to significant enhancements in convergence speed and performance.

\subsection{ Performance Related to the partitions}

We study the effect of the dependency graph of robots and the partition of rewards on the robots' overall performance. The following results will show that by delicately clustering the rewards of strongly related robots, the robots effectively learn a better policy.

In the aspect of performance, maximizing the number of partitions and the number of mixer inputs can reach the highest performance. However, the trade-off is increasing the computational time. In some machines with limited computational resources, the highest performance may not be their first concern. Therefore, we are also interested in understanding how the size of the partition affects the performance of reinforcement learning robots. In Fig. \ref{fig:PP}, we tested Loc-FACMAC with two different partitions: (1) 5 partitions, \{\{1,2\},\{3,4\},\{5,6\},\{7,8,9\}, \{10,11,12\}\}, (2) 2 partitions, \{\{1,2,3,4,5,6\},\{7,8,9,10,11,12\}\}, and 1 partition. The result matches our expectation that if the number of partitions increases, the higher performance can be.

\section{Conclusions}
\label{sec:conclusions}

In this paper, we introduce Loc-FACMAC, a novel MARL method that combines reward locality with the actor-critic framework to enhance robot performance and learning efficiency. We validate Loc-FACMAC's effectiveness across two tasks, where it outperforms existing methods. Our findings also highlight the correlation between framework structure, maximum reward, and convergence speed. Future research will address challenges such as constructing dependency graphs without prior knowledge and re-evaluating the Individual-Global-Max (IGM) assumption's applicability.  Addressing these challenges requires exploring constrained optimization in competitive settings and developing methods to model dynamic robot dependencies without prior assumptions.

\bibliographystyle{IEEEtran}
\bibliography{IEEEabrv, sample}  

\begin{thebibliography}{10}
\providecommand{\url}[1]{#1}
\csname url@rmstyle\endcsname
\providecommand{\newblock}{\relax}
\providecommand{\bibinfo}[2]{#2}
\providecommand\BIBentrySTDinterwordspacing{\spaceskip=0pt\relax}
\providecommand\BIBentryALTinterwordstretchfactor{4}
\providecommand\BIBentryALTinterwordspacing{\spaceskip=\fontdimen2\font plus
\BIBentryALTinterwordstretchfactor\fontdimen3\font minus \fontdimen4\font\relax}
\providecommand\BIBforeignlanguage[2]{{%
\expandafter\ifx\csname l@#1\endcsname\relax
\typeout{** WARNING: IEEEtran.bst: No hyphenation pattern has been}%
\typeout{** loaded for the language `#1'. Using the pattern for}%
\typeout{** the default language instead.}%
\else
\language=\csname l@#1\endcsname
\fi
#2}}

\bibitem{foerster2017counterfactual}
J.~Foerster, G.~Farquhar, T.~Afouras, N.~Nardelli, and S.~Whiteson, ``Counterfactual multi-agent policy gradients,'' 2017.

\bibitem{tan1993multi}
M.~Tan, ``Multi-agent reinforcement learning: Independent vs. cooperative agents,'' in \emph{Proceedings of the tenth international conference on machine learning}, 1993, pp. 330--337.

\bibitem{jiang2022multi}
Q.~Jiang, M.~Qin, S.~Shi, W.~Sun, and B.~Zheng, ``Multi-agent reinforcement learning for traffic signal control through universal communication method,'' \emph{arXiv preprint arXiv:2204.12190}, 2022.

\bibitem{chu2019multi}
T.~Chu, J.~Wang, L.~Codec{\`a}, and Z.~Li, ``Multi-agent deep reinforcement learning for large-scale traffic signal control,'' \emph{IEEE Transactions on Intelligent Transportation Systems}, vol.~21, no.~3, pp. 1086--1095, 2019.

\bibitem{zhang2022ai}
T.~Zhang, A.~Williams, S.~Phade, S.~Srinivasa, Y.~Zhang, P.~Gupta, Y.~Bengio, and S.~Zheng, ``Ai for global climate cooperation: Modeling global climate negotiations, agreements, and long-term cooperation in rice-n,'' \emph{arXiv preprint arXiv:2208.07004}, 2022.

\bibitem{he2016faster}
X.~He, H.~Dai, and P.~Ning, ``Faster learning and adaptation in security games by exploiting information asymmetry,'' \emph{IEEE Transactions on Signal Processing}, vol.~64, no.~13, pp. 3429--3443, 2016.

\bibitem{arul2022multi}
S.~H. Arul, A.~S. Bedi, and D.~Manocha, ``Multi robot collision avoidance by learning whom to communicate,'' \emph{arXiv preprint arXiv:2209.06415}, 2022.

\bibitem{qie2019joint}
H.~Qie, D.~Shi, T.~Shen, X.~Xu, Y.~Li, and L.~Wang, ``Joint optimization of multi-uav target assignment and path planning based on multi-agent reinforcement learning,'' \emph{IEEE access}, vol.~7, pp. 146\,264--146\,272, 2019.

\bibitem{vorotnikov2018multi}
S.~Vorotnikov, K.~Ermishin, A.~Nazarova, and A.~Yuschenko, ``Multi-agent robotic systems in collaborative robotics,'' in \emph{Interactive Collaborative Robotics: Third International Conference, ICR 2018, Leipzig, Germany, September 18--22, 2018, Proceedings 3}.\hskip 1em plus 0.5em minus 0.4em\relax Springer, 2018, pp. 270--279.

\bibitem{jimenez2018decentralized}
A.~C. Jim{\'e}nez, V.~Garc{\'\i}a-D{\'\i}az, and S.~Bola{\~n}os, ``A decentralized framework for multi-agent robotic systems,'' \emph{Sensors}, vol.~18, no.~2, p. 417, 2018.

\bibitem{ji2022traversing}
T.~Ji, R.~Dong, and K.~Driggs-Campbell, ``Traversing supervisor problem: An approximately optimal approach to multi-robot assistance,'' \emph{arXiv preprint arXiv:2205.01768}, 2022.

\bibitem{ota2006multi}
J.~Ota, ``Multi-agent robot systems as distributed autonomous systems,'' \emph{Advanced engineering informatics}, vol.~20, no.~1, pp. 59--70, 2006.

\bibitem{reis2023coordinationmachinelearningmultirobot}
\BIBentryALTinterwordspacing
L.~P. Reis, ``Coordination and machine learning in multi-robot systems: Applications in robotic soccer,'' 2023. [Online]. Available: \url{https://arxiv.org/abs/2312.16273}
\BIBentrySTDinterwordspacing

\bibitem{christianos2020shared}
\BIBentryALTinterwordspacing
F.~Christianos, L.~Sch\"{a}fer, and S.~Albrecht, ``Shared experience actor-critic for multi-agent reinforcement learning,'' in \emph{Advances in Neural Information Processing Systems}, H.~Larochelle, M.~Ranzato, R.~Hadsell, M.~F. Balcan, and H.~Lin, Eds., vol.~33.\hskip 1em plus 0.5em minus 0.4em\relax Curran Associates, Inc., 2020, pp. 10\,707--10\,717. [Online]. Available: \url{https://proceedings.neurips.cc/paper/2020/file/7967cc8e3ab559e68cc944c44b1cf3e8-Paper.pdf}
\BIBentrySTDinterwordspacing

\bibitem{papoudakis2021benchmarking}
\BIBentryALTinterwordspacing
G.~Papoudakis, F.~Christianos, L.~Schäfer, and S.~V. Albrecht, ``Benchmarking multi-agent deep reinforcement learning algorithms in cooperative tasks,'' in \emph{Proceedings of the Neural Information Processing Systems Track on Datasets and Benchmarks (NeurIPS)}, 2021. [Online]. Available: \url{http://arxiv.org/abs/2006.07869}
\BIBentrySTDinterwordspacing

\bibitem{watkins1992q}
C.~J. Watkins and P.~Dayan, ``Q-learning,'' \emph{Machine learning}, vol.~8, no.~3, pp. 279--292, 1992.

\bibitem{sunehag2017value}
P.~Sunehag, G.~Lever, A.~Gruslys, W.~M. Czarnecki, V.~Zambaldi, M.~Jaderberg, M.~Lanctot, N.~Sonnerat, J.~Z. Leibo, K.~Tuyls, \emph{et~al.}, ``Value-decomposition networks for cooperative multi-agent learning,'' \emph{arXiv preprint arXiv:1706.05296}, 2017.

\bibitem{pmlr-v80-rashid18a}
\BIBentryALTinterwordspacing
T.~Rashid, M.~Samvelyan, C.~Schroeder, G.~Farquhar, J.~Foerster, and S.~Whiteson, ``{QMIX}: Monotonic value function factorisation for deep multi-agent reinforcement learning,'' in \emph{Proceedings of the 35th International Conference on Machine Learning}, ser. Proceedings of Machine Learning Research, J.~Dy and A.~Krause, Eds., vol.~80.\hskip 1em plus 0.5em minus 0.4em\relax PMLR, 10--15 Jul 2018, pp. 4295--4304. [Online]. Available: \url{https://proceedings.mlr.press/v80/rashid18a.html}
\BIBentrySTDinterwordspacing

\bibitem{rashid2020weighted}
T.~Rashid, G.~Farquhar, B.~Peng, and S.~Whiteson, ``Weighted qmix: Expanding monotonic value function factorisation for deep multi-agent reinforcement learning,'' \emph{Advances in neural information processing systems}, vol.~33, pp. 10\,199--10\,210, 2020.

\bibitem{son2019qtran}
K.~Son, D.~Kim, W.~J. Kang, D.~E. Hostallero, and Y.~Yi, ``Qtran: Learning to factorize with transformation for cooperative multi-agent reinforcement learning,'' in \emph{International conference on machine learning}.\hskip 1em plus 0.5em minus 0.4em\relax PMLR, 2019, pp. 5887--5896.

\bibitem{kortvelesy2022qgnn}
R.~Kortvelesy and A.~Prorok, ``Qgnn: Value function factorisation with graph neural networks,'' 2022.

\bibitem{9533636}
Z.~Xu, D.~Li, Y.~Bai, and G.~Fan, ``Mmd-mix: Value function factorisation with maximum mean discrepancy for cooperative multi-agent reinforcement learning,'' in \emph{2021 International Joint Conference on Neural Networks (IJCNN)}, 2021, pp. 1--7.

\bibitem{konda1999actor}
V.~Konda and J.~Tsitsiklis, ``Actor-critic algorithms,'' \emph{Advances in neural information processing systems}, vol.~12, 1999.

\bibitem{NEURIPS2021_65b9eea6}
\BIBentryALTinterwordspacing
B.~Peng, T.~Rashid, C.~Schroeder~de Witt, P.-A. Kamienny, P.~Torr, W.~Boehmer, and S.~Whiteson, ``Facmac: Factored multi-agent centralised policy gradients,'' in \emph{Advances in Neural Information Processing Systems}, M.~Ranzato, A.~Beygelzimer, Y.~Dauphin, P.~Liang, and J.~W. Vaughan, Eds., vol.~34.\hskip 1em plus 0.5em minus 0.4em\relax Curran Associates, Inc., 2021, pp. 12\,208--12\,221. [Online]. Available: \url{https://proceedings.neurips.cc/paper/2021/file/65b9eea6e1cc6bb9f0cd2a47751a186f-Paper.pdf}
\BIBentrySTDinterwordspacing

\bibitem{wang2020r}
R.~E. Wang, M.~Everett, and J.~P. How, ``R-maddpg for partially observable environments and limited communication,'' \emph{arXiv preprint arXiv:2002.06684}, 2020.

\bibitem{samvelyan2019starcraft}
M.~Samvelyan, T.~Rashid, C.~S. De~Witt, G.~Farquhar, N.~Nardelli, T.~G. Rudner, C.-M. Hung, P.~H. Torr, J.~Foerster, and S.~Whiteson, ``The starcraft multi-agent challenge,'' \emph{arXiv preprint arXiv:1902.04043}, 2019.

\bibitem{lowe2017multi}
R.~Lowe, Y.~I. Wu, A.~Tamar, J.~Harb, O.~Pieter~Abbeel, and I.~Mordatch, ``Multi-agent actor-critic for mixed cooperative-competitive environments,'' \emph{Advances in neural information processing systems}, vol.~30, 2017.

\bibitem{kolling2008multi}
A.~Kolling and S.~Carpin, ``Multi-robot surveillance: an improved algorithm for the graph-clear problem,'' in \emph{2008 IEEE International Conference on Robotics and Automation}.\hskip 1em plus 0.5em minus 0.4em\relax IEEE, 2008, pp. 2360--2365.

\bibitem{zohar2022locality}
R.~Zohar, S.~Mannor, G.~Tennenholtz, , and and, ``Locality matters: A scalable value decomposition approach for cooperative multi-agent reinforcement learning,'' in \emph{Proceedings of the AAAI Conference on Artificial Intelligence}, vol.~36, 2022, pp. 9278--9285.

\bibitem{hao2023boosting}
\BIBentryALTinterwordspacing
J.~HAO, X.~Hao, H.~Mao, W.~Wang, Y.~Yang, D.~Li, Y.~ZHENG, and Z.~Wang, ``Boosting multiagent reinforcement learning via permutation invariant and permutation equivariant networks,'' in \emph{The Eleventh International Conference on Learning Representations}, 2023. [Online]. Available: \url{https://openreview.net/forum?id=OxNQXyZK-K8}
\BIBentrySTDinterwordspacing

\bibitem{liu2019multiagent}
Y.~Liu, W.~Wang, Y.~Hu, J.~Hao, X.~Chen, and Y.~Gao, ``Multi-agent game abstraction via graph attention neural network,'' 2019.

\bibitem{li2021deep}
S.~Li, J.~K. Gupta, P.~Morales, R.~Allen, and M.~J. Kochenderfer, ``Deep implicit coordination graphs for multi-agent reinforcement learning,'' 2021.

\bibitem{wang2022contextaware}
T.~Wang, L.~Zeng, W.~Dong, Q.~Yang, Y.~Yu, and C.~Zhang, ``Context-aware sparse deep coordination graphs,'' 2022.

\bibitem{böhmer2020deep}
W.~Böhmer, V.~Kurin, and S.~Whiteson, ``Deep coordination graphs,'' 2020.

\end{thebibliography}

\end{document}